%% file: main.tex
\documentclass{article}
\usepackage{spconf,amsmath,graphicx}

\title{An Information-Theoretic Approach to Transferability in Task Transfer Learning}
\name{Yajie Bao$^{1\star }$~  Yang Li$^{1\star}$\thanks{$^\star$ Joint-first authors}~  Shao-Lun Huang$^{1}$~ Lin Zhang$^{1}$~  Lizhong Zheng$^{2}$~ Amir Zamir$^{3,4}$~ Leonidas Guibas$^{3}$}

\address{\begin{tabular}{cc}
$^{1}$ Tsinghua-Berkeley Shenzhen Institute &
    $^{2}$ Massachusetts Institute of Technology\\
    $^{3}$ Stanford University  & $^{4}$ University of California, Berkeley 
 \end{tabular}
}
\input{math_commands.tex}

\usepackage[inline]{enumitem}

\usepackage{hyperref}
\usepackage{url} 
\usepackage{booktabs}       
\usepackage{amsfonts}       
\usepackage{nicefrac}       
\usepackage{microtype}      
\usepackage{amsthm}
\usepackage{amssymb} 
\usepackage{amsmath}
\usepackage{graphicx}
\usepackage{caption}
\usepackage{comment} 
\usepackage{multirow}
\usepackage{color}
\usepackage[usenames,dvipsnames]{xcolor}
\usepackage{colortbl}
\usepackage[title]{appendix} 
\usepackage{pgfplots}

\usepackage{setspace}
\pgfmathdeclarefunction{gauss}{2}{%
  \pgfmathparse{1/(#2*sqrt(2*pi))*exp(-((x-#1)^2)/(2*#2^2))}%
} 
\definecolor{sred}{RGB}{221,160,221}
\definecolor{sblue}{RGB}{0,255,255}

\newtheorem{lemma}{Lemma}
\newtheorem{thm}{Theorem}

\newtheorem{definition}{Definition}
\usepackage{soul}
\usepackage{algorithm}
\usepackage{algorithmic}

\usepackage{framed}
\colorlet{shadecolor}{blue!20}

\def\squeze{\vspace{-0.7em}}

 \usepackage{xpatch}

 \usepackage[compact]{titlesec} 
\makeatletter

\g@addto@macro\normalsize{%
  \setlength\abovedisplayskip{2pt}
  \setlength\belowdisplayskip{0pt}
 \setlength{\textfloatsep}{1\baselineskip plus 0.2\baselineskip minus 0.5\baselineskip}

}

 \xpatchcmd{\footnotesize}{{9pt}{9pt}}{}{\patchfailed}
\begin{document}
%
\maketitle
\begin{abstract}
Task transfer learning is a popular technique in image processing applications that uses pre-trained models to reduce the supervision cost of related tasks. An important question is to determine task transferability, i.e. given  a common input domain, estimating to what extent  representations learned from a source task can help in learning a target task. Typically, transferability is either measured experimentally or inferred through task relatedness, which is often defined without a clear operational meaning.  In this paper, we present a novel metric, {\em $H$-score}, an easily-computable evaluation function that estimates the performance of transferred representations from one task to another in classification problems using statistical and information theoretic principles.  Experiments on real image data show that our metric is not only consistent with the empirical transferability measurement, but also useful to practitioners in applications such as source model selection and task transfer curriculum learning. 
\end{abstract}
\input{intro}
\input{hscore}

\input{transferability}
\input{experiments_simplified}
\input{conclusion} 
\section*{Acknowledgment}\addcontentsline{toc}{section}{Acknowledgment}

{\fontsize{8pt}{0pt}\footnotesize{\setlength{\baselineskip}{0pt}
This research is funded by Natural Science Foundation of China 61807021,  Shenzhen Science and Technology Research and Development Funds JCYJ20170818094022586, Innovation and Entrepreneurship Project for Overseas High-Level Talents of Shenzhen KQJSCX2018032714403783, a grant from the Toyota-Stanford Center for AI Research, a Vannevar Bush Faculty Fellowship, and NSF grant DMS-1546206. Toyota Research Institute (``TRI'')  provided funds to assist the authors with their research but this article solely reflects the opinions and conclusions of its authors and not TRI or any other Toyota entity.} }
\normalsize

\bibliographystyle{IEEEbib}
\bibliography{main} 
\vfill\eject

\input{supplementary}  
\end{document}

%% file: math_commands.tex
\usepackage{amsmath,amsfonts,bm}

\def\T{\mathcal{T}}

\def\eqref#1{equation~\ref{#1}}

\def\1{\bm{1}}

\DeclareMathAlphabet{\mathsfit}{\encodingdefault}{\sfdefault}{m}{sl}
\SetMathAlphabet{\mathsfit}{bold}{\encodingdefault}{\sfdefault}{bx}{n}

\newcommand{\E}{\mathbb{E}}

\newcommand{\R}{\mathbb{R}}

 \def\X{\mathcal{X}}
 
 \def\Y{\mathcal{Y}}
 
 \def\H{\mathcal{H}}
 
 \def\cov{\textup{cov}} 
 \def\opt{\textup{opt}} 
 \def\tr{\textup{tr}}
 \def\B{\tilde{B}}
 
\newcommand{\squeze}{\vspace{-1.5mm}}
\newcommand{\squezeup}{\vspace{-2.5mm}}

 \DeclareMathOperator*{\argmin}{arg\!\min}
 \DeclareMathOperator*{\argmax}{\arg\!\max} 

%% file: intro.tex
\section{Introduction}
\begin{figure}[t]
\centering
 \includegraphics[width=0.47\textwidth]{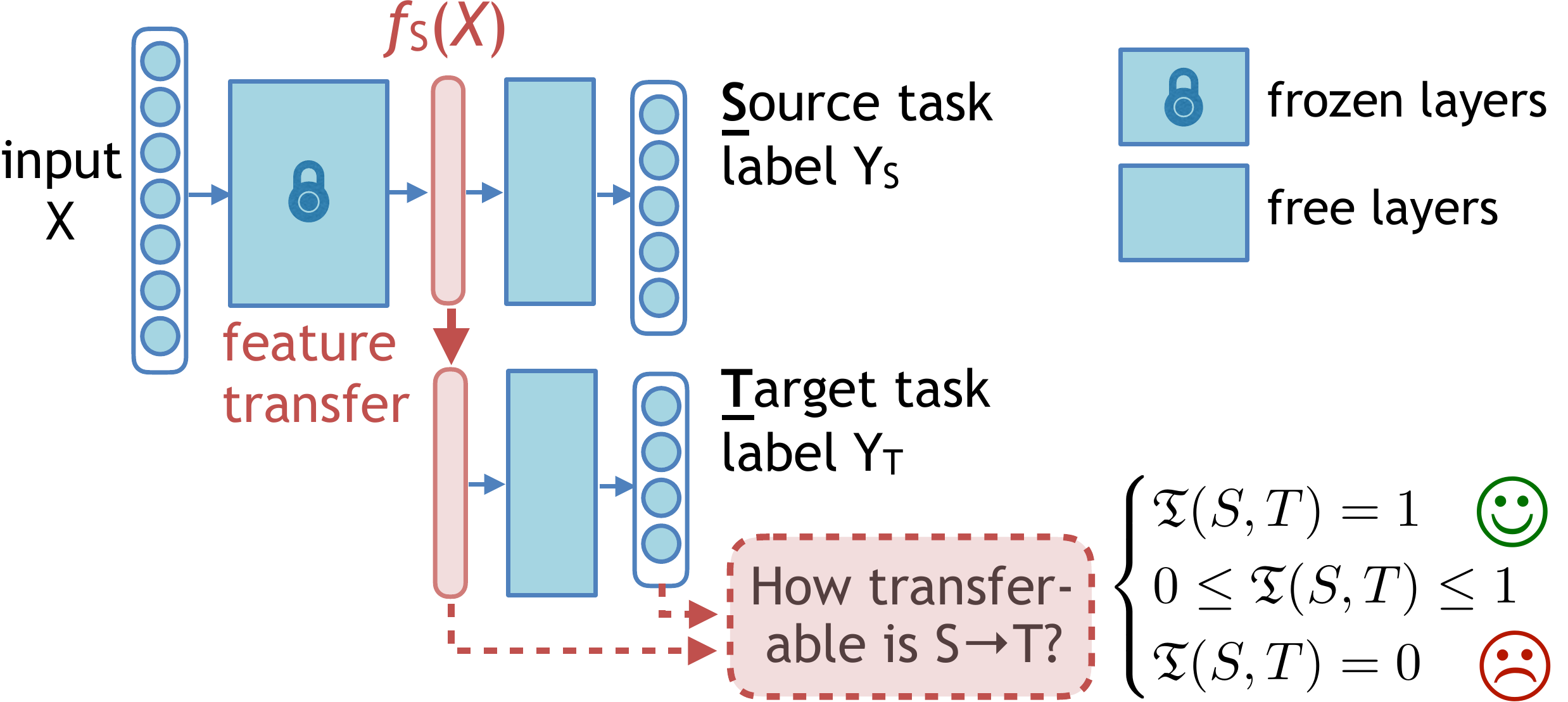}
\caption{\label{fig:tl-network} A generic model of task transfer learning. The proposed transferability metric $\mathfrak{T}(S,T)$ can predict the target task's performance without training the task transfer network.} 
\end{figure}

 {\em Transfer learning} is a learning paradigm that exploits the relatedness between different learning tasks in order to gain certain benefits, e.g. reducing the demand for supervision  (\cite{pratt1993}).  In {\em task transfer learning}, we assume that the input domain of the different tasks are the same. Then for a target task $\mathcal{T}_T$, instead of learning a model from scratch, we can initialize the parameters from a previously trained model for some related source task $\mathcal{T}_S$ (Figure \ref{fig:tl-network}). For example, deep convolutional neural networks trained for the ImageNet classification task have been used as the source network in transfer learning for target tasks with fewer labeled data \cite{donahue2014decaf}, such as medical image analysis \cite{7318461} and  structural damage recognition in buildings \cite{gaodeep}. 

An imperative question in task transfer learning is {\em transferability}, i.e. {\em when a transfer may work} and {\em to what extent}. Given a metric capable of efficiently and accurately measuring transferability across arbitrary tasks, the problem of task transfer learning, to a large extent, is simplified to search 
procedures over potential transfer sources and targets as quantified by the metric. Traditionally, transferability is measured purely empirically using model loss or accuracy on the validation set (\cite{yosinski2014transferable,zamir2018taskonomy,conneau2017supervised}).   
There have  been theoretical studies that focus on {\em task relatedness} (\cite{baxter2000model,maurer2009transfer,Pentina2014A,ben2003exploiting}). 
However, they either cannot be computed explicitly from data or do not directly explain task transfer performance. In this study, {\em we aim to estimate transferability analytically, directly from the training data}.

We quantify the transferability of feature representations  across tasks via an approach grounded in statistics and information theory. 
The key idea of our method is to show that the expected log-loss  of using a feature of the input data to predict the label of a given task under the probabilistic model can be characterized by an analytically expression, which we refer as the \textit{H-score}  of the feature.  
H-score is particularly useful to quantify feature transferability among tasks. Using this idea, we define {\em task transferability} as the normalized H-score of the optimal source task feature with respect to the target task.  

As we demonstrate in this paper, the advantage of our transferability metric is threefold. \begin{enumerate*}[label=(\roman*)]
\item it is theoretically driven and has a strong operational meaning rooted in statistics and information theory;
\item it can be computed directly and efficiently from the input data, with fewer samples than those needed for empirical learning;
\item it can be shown to be strongly consistent with empirical transferability measurements. 
\end{enumerate*}  

%% file: hscore.tex
 \section{Measuring Feature Effectiveness}
 \label{sec:measuringEffectiveness}
Let $\X$ and $\Y$ denote the input and output space respectively. Denote the transferred feature representation by $f:\X\rightarrow \R^k$.  For a classification task,  let   $h_f:\X\times\Y \rightarrow [0,1]^{|\Y|}$ be a predictor function 
with the log-loss function $L(f(x),y)$ for a given $(x,y)$ sample.  
 The traditional machine learning approach uses stochastic gradient descent to minimize $L(h) = \E_{X,Y}[ L(f(x),y )] $.  
 We will show that the optimal log loss when $f$ is given can be characterized analytically using concepts in  information theory and statistics.   
\begin{definition}  
The {\em Divergence Transition Matrix (DTM)}  of discrete random variables $X$ and $Y$ is a $|\Y|$ by $|\X|$ matrix $\hat{B}$ with entries $\hat{B}_{y,x} = \frac{P_{XY}(x,y)}{\sqrt{P_X(x)}\sqrt{P_Y(y)}} - \sqrt{P_Y(y)}\sqrt{P_X(x)}$ for all $x\in \X$ and $y\in \Y$. 
\end{definition} 
  Given $m$ training examples $\{(x^{(i)}, y^{(i)})\}_{i=1}^m$, $L(h)$ can be written as 
\begin{align*}
L(f,\theta) =& -\frac{1}{m}\sum_{i=1}^m \sum_{k=1}^{|\Y|}  1\{y^{(i)} = k\} \log  \frac{ e^{-\theta^T_k f(x^{(i)})}}{\sum_{j=1}^{|\Y|} e^{-\theta^T_j f(x^{(i)})}} 
\end{align*} 
Using concepts in Euclidean information geometry, it is shown in \cite{acejournal} that under a local assumption, for a given feature dimension $k$, 
\begin{equation}
\argmin_{f,\theta}L(f,\theta) = 
\argmin_{\Psi\in\R^{|\X|\times k}, \Phi\in\R^{|\Y|\times k}} \frac{1}{2}\|\tilde{B}-\Psi \Phi^{\mathrm{T}}\|_{F}^{2} + o(\epsilon^{2})\label{op:s2}
\end{equation}
Let $\phi(x)$ represent row vectors of $\Phi$ for any $x\in \X$.
By defining a one-to-one mapping   $f(x)\leftrightarrow \phi(x)$ such that 
$\phi(x)= \sqrt{P_X(x)}f(x)$, 
 Eq.(\ref{op:s2}) reveals a close connection between the optimal log-loss and the modal decomposition of $\tilde{B}$. In consequence, it is reasonable to measure the classification performance with $\|\tilde{B}-\Psi \Phi^{\mathrm{T}}\|_{F}^{2}$ given $f(X)$. i.e.     Since $\Phi$ is fixed, we can find the optimal $\Psi$ , $\Psi^*$  by taking the derivative of the objective function with respect to $\Psi$:  
\begin{equation}
\Psi ^{*}=\tilde{B}\Phi(\Phi ^{\mathrm{T}}\Phi)^{-1}  \label{eq:s3}
\end{equation}

Substituting (\ref{eq:s3}) in the Objective of (\ref{op:s2}), we can derive the following close-form solution for the log loss:
\begin{equation}
\|\tilde{B}\|_{F}^{2}-\|\tilde{B}\Phi (\Phi ^{\mathrm{T}}\Phi )^{-\frac{1}{2}}\|_{F}^{2}  \label{eq:s7}
\end{equation}

The first term in (\ref{eq:s7}) does not depend on $f(X)$, therefore it is sufficient to use the second term to estimate classification performance with transferred feature $f$. We can further rewrite $||\tilde{B}\Phi(\Phi^T\Phi)^{-\frac{1}{2}}||_F^2$ as follows and denote it as the {\em H-score}.
\begin{definition}\label{def:hscore} Given data matrix $X\in \R^{m\times d}$ and label $Y$, let $f(X)$  be a $k$-dim, zero-mean feature function. The H-Score of $f$ with respect to a task with joint probability $P_{YX}$  is: 
\begin{equation}\label{eq:hscore}
    \H(f) =   \textup{tr}(\cov(f(X))^{-1}\cov( \E_{P_{X|Y}}[f(X)|Y]))
\end{equation}

\end{definition}
The derivation of (\ref{eq:hscore}) can be found in Section 1 of the Supplementary Material\footnote{Supplementary materials, data and code are available at \href{http://www.yangli-feasibility.com/home/ttl.html}{http://yangli-feasibility.com/home/ttl.html}}.  This formulation  can be intuitively interpreted from a nearest neighbor perspective. i.e.  a high H-score implies  the  inter-class variance $\cov(\E_{P_{X|Y}}[f(X)|Y])$ of $f$ is large, while feature redundancy $\tr(\cov(f(X)))$ is small.  
 Comparing to finding the optimal log-loss through gradient descent, H-score can be computed analytically and only requires estimating the conditional expectation  $\E[f(X)|Y]$  from sample data.  Moreover,  $\H(f) $ has an operational meaning that   characterizes the asymptotic error probability of using $f(X)$ to estimate $Y$ in the hypothesis testing context. (See Section 2 in the Supplementary Material  for  details).  

The upper bound of $\H(f)$ is obvious from its first definition: $\max_{\Phi}||B\Phi(\Phi^T\Phi)^{-\frac{1}{2}}||_F^2 = ||\tilde{B}||_F^2$. 
We call features that achieve this bound the {\em minimum error probability features} for a given task.

%% file: transferability.tex
\section{Transferability } 
\label{sec:transferability}
Next, we apply  H-score to efficiently measure the effectiveness of   task transfer learning. We will use subscripts $S$ and $T$ to distinguish variables for the source and the target tasks.  


\begin{definition}[Task transferability] Given source task $\mathcal{T}_S$, target task $\mathcal{T}_T$ and pre-trained source feature $f_S(x)$, the  {\em transferability from $\mathcal{T}_S$ to $\mathcal{T}_T$ } is  
$ \mathfrak{T}(S,T)\triangleq\frac{\H_{T}(f_S )}{\H_{T}(f_{T_{\opt}})} 
$, where   $f_{T_{\opt}}(x)$ is the minimum error probability feature of  the target task.
\end{definition}  

This definition   implies $0\leq \mathfrak{T}(S,T)\leq 1$. 
With a known $f$, computing H-score from $m$ sample data  only takes $O(mk^2)$ time,  where $k$ is the dimension of $f(x)$ for $k< m$. The majority of the computation time is spent on computing the sample covariance matrix $\textup{cov}(f(X))$. 
 
The remaining question is how to obtain $\H_T(f_{T_{opt}})$ efficiently. This question has been addressed in \cite{huang2017information}, which  shows that    $||\tilde{B}_{T}||_F^2 = \E[f(X)^\mathrm{T} g(Y)]$, where $f$  and $g$ are the solutions of the HGR-Maximum Correlation problem. 
\begin{align}\label{eq:hgr} 
  \rho(X;Y)= \sup_{\small\begin{matrix}f:\X\rightarrow\R^k,g:\Y\rightarrow\R^k\\\E[f(X)]=\E[g(Y)] =   0\\\E[f(X)f(X)^T]=I \end{matrix}} ~ & \E[f(X)^Tg(Y)] 
  \end{align} 
Eq. (6) can be solved efficiently using the Alternating Conditional Expectation (ACE) algorithm \cite{breiman1985estimating} for discrete $\X$, or using the neural network approach based on Generalized Maximal HGR Correlation \cite{wang2019} for a generic $\X$.
The sample complexity of ACE is only $1/k$ of the complexity of estimating $P_{YX}$ directly \cite{makur2015efficient}. This result also applies to the Generalized HGR problem  due to their theoretical equivalence. 

A common technique in task transfer learning is fine-tuning, which adds before the target classifier additional free layers, whose parameters are optimized with respect to the target label.  For the operational meaning of transferability to hold exactly,  we require the fine tuning layers consist of only linear transformations. 
 Nevertheless, later we will demonstrate empirically that this transferability metric can still be used for comparing the {\em relative} task transferability with fine-tuning. 
In many cases though, the computation of $H_T(f_{opt})$ can  even be skipped entirely, such as the problem below:


\begin{definition}[Source task selection]
Given $N$ source tasks $\mathcal{T}_{S_1},\dots, \mathcal{T}_{S_N}$ with labels $Y_{S_1},\dots, Y_{S_N}$ and a target task $\mathcal{T}_T$ with label $Y_T$. Let  $f_{S_1},\dots, f_{S_N}$  be optimal representations for the source tasks. Find the  source task $\mathcal{T}_{S_i}$ that maximizes the testing accuracy of predicting   $Y_T$ with feature $f_{S_i}$.
\end{definition}

We can solve this problem by selecting the source task with the largest transferability to $\mathcal{T}_T$.  In fact, we only need to compute the numerator in the transferability definition since the denominator is the same for all source tasks, i.e. 
$\argmax_i \mathfrak{T}(S_i,T)
=\argmax_i \mathcal{H}(f_{S_i})
$. 

%% file: experiments_simplified.tex
\section{Experiments}
\label{sec:experiments}
In this section, we present validation results and potential application of our transferability metric on real image data. (For implementation details, see Section 3 of the Supplementary Material.) 
 
\subsection{Validation of transfer performance}  
\begin{figure}[htp]
\centering
\renewcommand{\arraystretch}{0.8}
\begin{tabular}{cc }
\includegraphics[height=0.12\textheight]{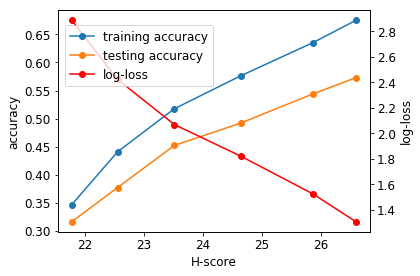} &
\includegraphics[height=0.12\textheight]{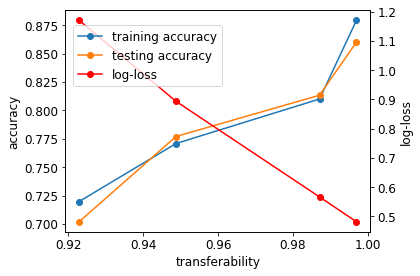} \\
(a) & (b)  
\end{tabular}\vspace{-0.8em}
\caption{\label{fig:cifar}H-score and transferability vs. the empirical transfer performance measured by log-loss, training and testing accuracy. a.) Performance of ImageNet-1000 features from layers 4a-4f for Cifar-100 classification.  
b.): Transferability from ImageNet-1000 to 4 different target tasks based on Cifar-100.  }
\label{fig:logloss}
\end{figure}

We validate H-score and transferability definitions in a transfer learning problem from ImageNet 1000-class classification (ImageNet-1000) to Cifar 100-class classification (Cifar-100).   Figure \ref{fig:cifar}.a compares the H-score and empirical performance of transferring from five different layers (4a-4f) of the ResNet-$50$ model pretrained on ImageNet1000.  
As H-score increases,  log-loss of the target network decreases almost linearly while the training and testing accuracy increase, which validates the relationship between the expected log-loss and H-score. The training and testing accuracy are also positively correlated with H-score.    
It also shows that H-score can be applied for selecting the most suitable layer for fine-tuning in transfer learning. 

 We further tested our transferability metric for selecting the best target task for a given source task. In particular, we constructed 4 target classification tasks with 3, 5, 10, and 20 object categories from the Cifar-100 dataset.  We then computed the task  transferability from ImageNet-1000 (using the feature representation of layer 4f) to the target tasks. 
 In Figure\ref{fig:logloss}.b, we observe a similar behavior as the H-score in the case of a single target task in Figure~\ref{fig:logloss}.a, showing that transferability can directly predict the  empirical transfer performance.  
\subsection{Task transfer for 3D scene understanding}  
Next, we apply our transferability metric to solve the source task selection problem among 8 image-based recognition tasks for 3D scene understanding using the Taskonomy dataset\cite{zamir2018taskonomy}. We also compared the task transferability ranking based on H-score with the ranking using {\em task affinity}, an empirical transferability metric proposed by \cite{zamir2018taskonomy} with non-linear fine tuning. 

For a fair comparison, we use the same trained encoders in \cite{zamir2018taskonomy} to extract source features with dimension $k=2048$. 
It's worth noting that, six of eight tasks have images as their output. To compute the transferability for these pixel-to-pixel tasks, we cluster the pixel values of the output images in the training data into a palette of $16$ colors and then compute the H-score of the source features with respect to each pixel. The transferability of the task is computed as the average of the H-scores over all pixels. For larger images, H-score can be evaluated on super pixels instead of pixels to improve efficiency.  
\begin{figure}[ht]
 {\bf Table. 1: } List of image scene understanding tasks
\centering 
\vspace{-1em}
\begin{tabular}{l p{5cm}}
\toprule
\small Classification tasks:   & \small Object Class., Scene Class. \\ 
\small Pixel-to-pixel tasks:    &  \small Keypint2D, Edge3D, Keypoint2D, Edge2D, Reshading, Depth \\
\bottomrule\\
\end{tabular}
 
\includegraphics[width=0.5\textwidth]{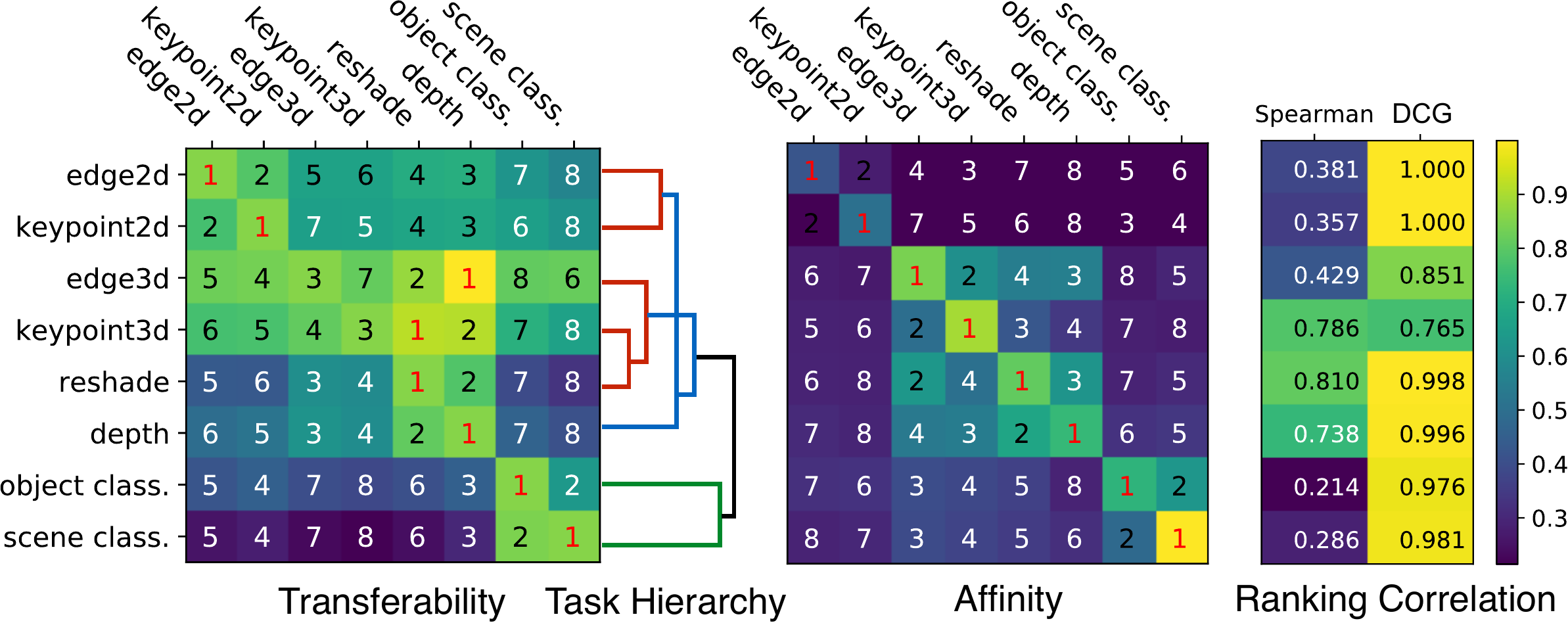} \vspace{-1em}
\caption{\label{fig:tasko-hscore} Ranking comparison between transferability and affinity score. }
\squeze
\end{figure}

{\noindent\bf Pairwise Transfer Results. } 
Source task ranking results using transferability and affinity are visualized side by side in Figure~\ref{fig:tasko-hscore}, with columns representing source tasks and rows representing target tasks. 
For classification tasks (the bottom two rows in the transferability matrix), the top two transferable source tasks are identical for both methods. 
Similar observations can be found in 2D pixel-to-pixel tasks (top two rows). 
A slightly larger difference between the two rankings can be found in 3D pixel-to-pixel tasks, especially 3D Occlusion Edges and 3D Keypoints. Though the top four ranked tasks of both methods are exactly the four 3D tasks. It could indicate that these low level vision tasks are closely related to each other so that the transferability among them are inherently ambiguous.  We also computed the ranking correlations between transferability and affinity using Spearman's R and Discounted Cumulative Gain (DCG). Both criterion show positive correlations for all target tasks. The correlation is especially strong with DCG as  higher ranking entities are given larger weights.

To show the task relatedness, we represent each task with a vector   consisting of H-scores of all the source tasks for the given task, then apply agglomerative clustering over the task vectors. As shown in the dendrogram in Figure~\ref{fig:tasko-hscore}, 2D tasks and most 3D tasks are grouped into different clusters, but on a higher level, all pixel-to-pixel tasks are considered one category compared to the classifications tasks. 
\begin{figure}[t]
\centering
 \includegraphics[width=0.49\textwidth]{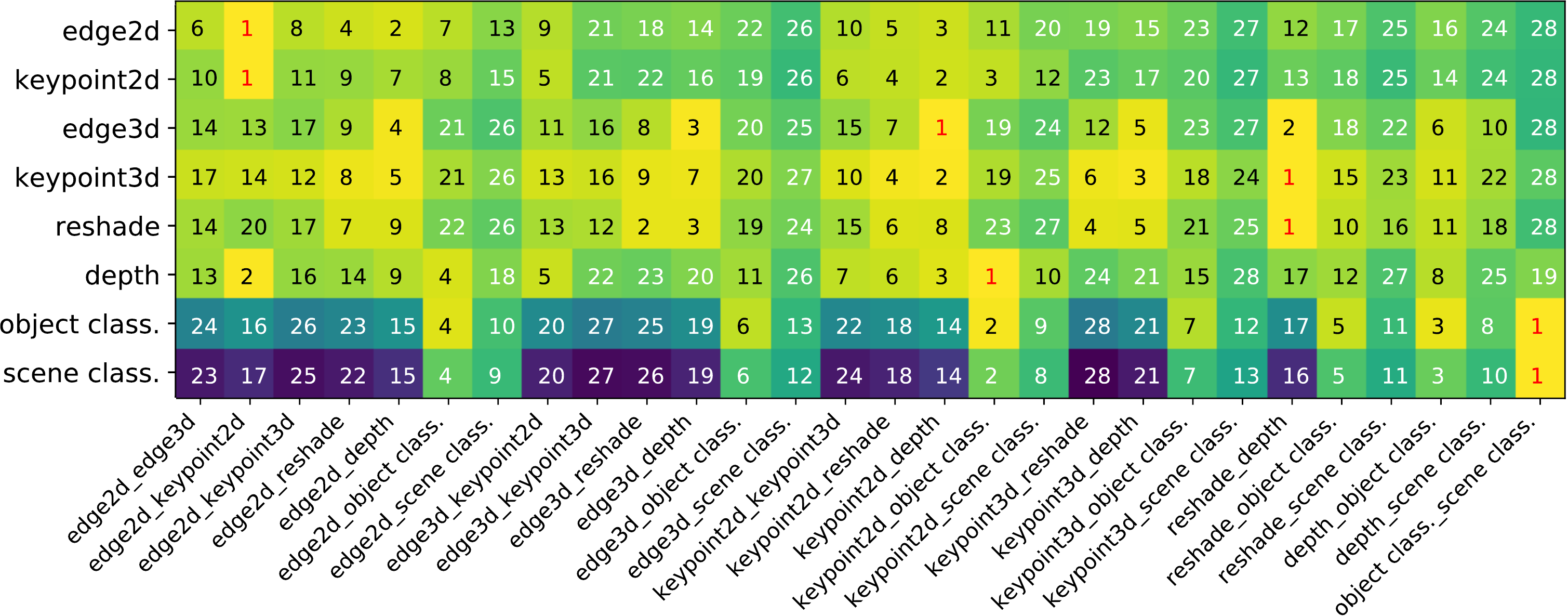}\vspace{-0.8em}
 \caption{\label{fig:order2ranking}Ranking of 2nd-order transferability for all tasks}
\includegraphics[width=0.42\textwidth]{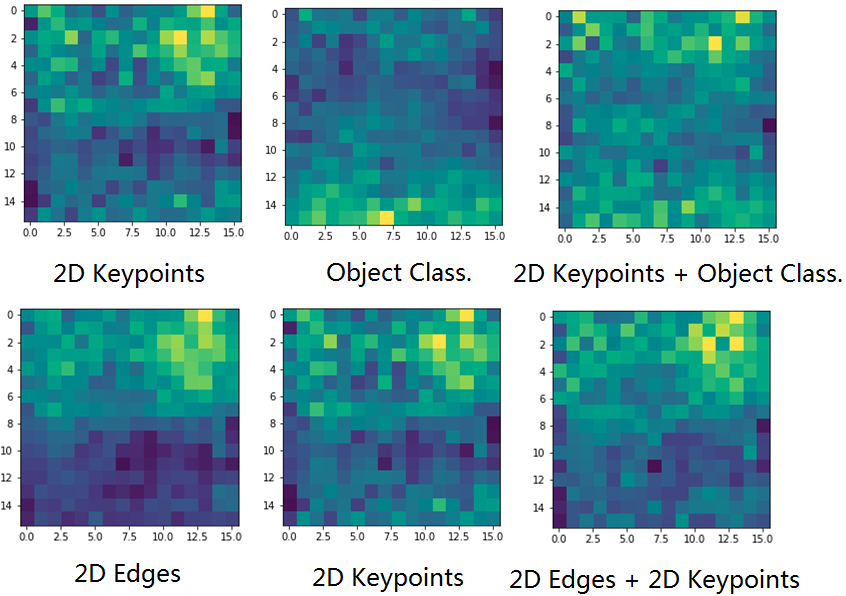}\vspace{-0.5em}
\caption{1st and 2nd order pixel-wise transferability to Depth.}
\label{fig:order2}  
\end{figure}

{\noindent\bf Higher Order Transfer.}  
 A common way for higher order transfer is to concatenate features  from multiple models in deep neural networks. Our transferability definition can be easily adapted to such problems. 
 Figure~\ref{fig:order2ranking} shows the ranking results of all combinations of source task pairs for each target task. For all tasks except for Edge3D and Depth,  the best seond-order source feature is the combination of the top two  tasks of the first-order ranking. We examine the exception in Figure ~\ref{fig:order2}, by visualizing the pixel-by-pixel H-scores of first and second order transfers to Depth using a heatmap (lighter color implies a higher H-score). Note that different source tasks can be good at predicting different parts of the image.   The top row shows the results of combining tasks with two different ``transferability patterns" while the bottom row shows those with similar patterns. Combining tasks with different transferability patterns has a more significant improvement to the overall performance of the target task. 

\subsection{Task transfer learning curriculum}
\begin{figure}[t] 
\centering
\includegraphics[height=0.10\textheight]{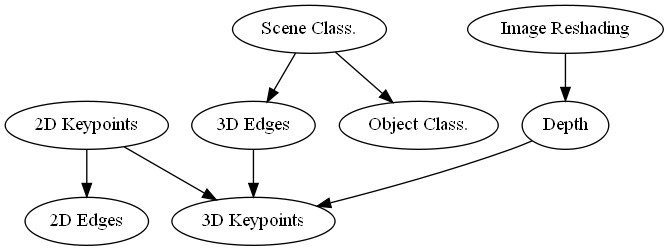}\vspace{-0.5em}
\caption{\label{fig:mst-taskonomy} Minimum spanning tree of task transferability. }
\centering  
\includegraphics[width=0.4\textwidth]{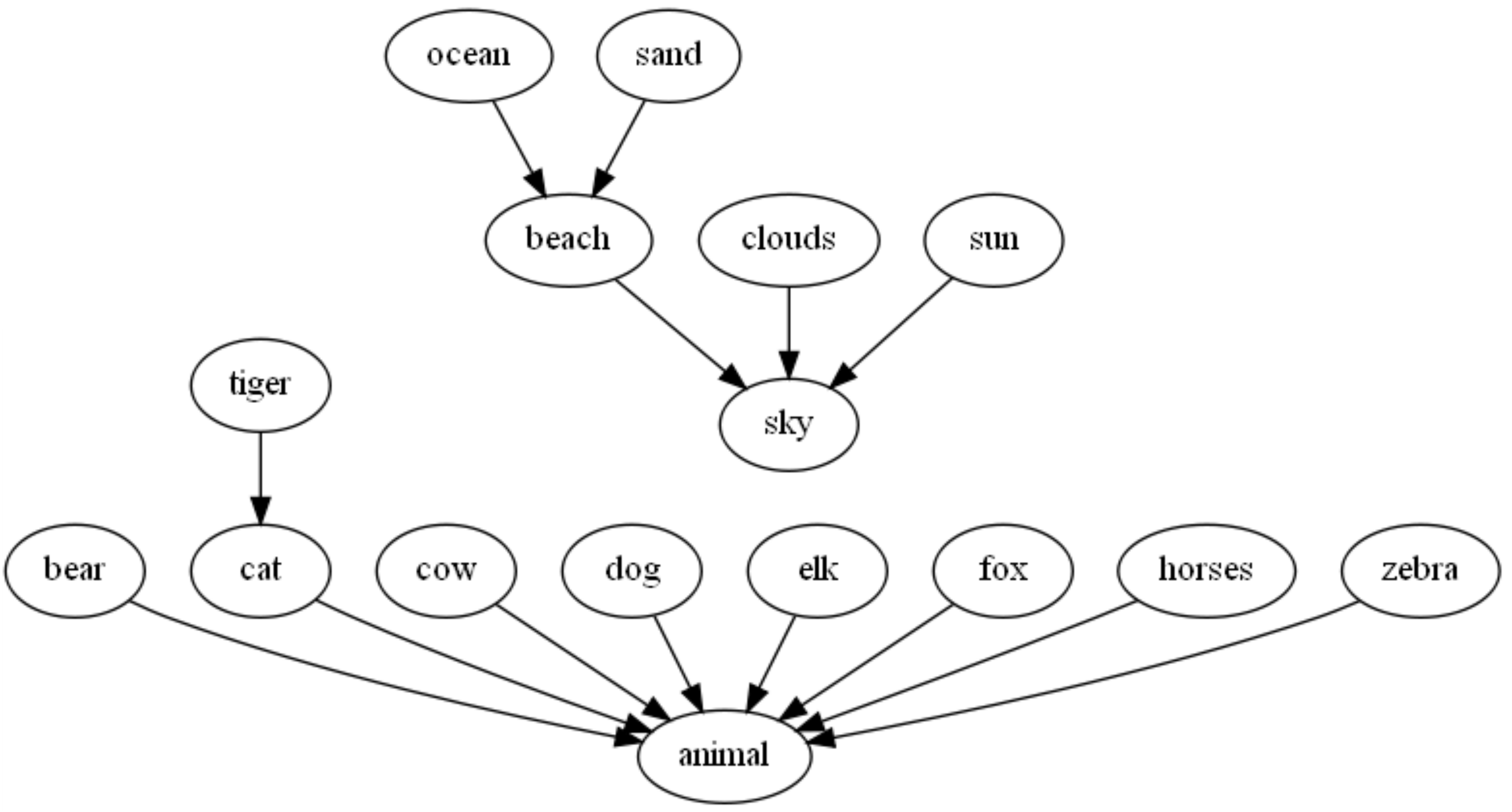}  \vspace{-0.8em}
\caption{\label{fig:minspantree} Minimum spanning trees of binary image classification tasks using the NUS-WIDE multi-label dataset. }
\end{figure}

A potential application of our transferability metric is developing an optimal task transfer curriculum, a directed acyclic graph over tasks that specifies the order  in which to obtain labeled data for  each task. For each task in the curriculum, an optimal feature representation can be learned using both its raw input and the representations of its parent tasks to improve training efficiency.   We use a heuristic based on the minimum spanning tree of a task graph, whose edge weights are inversely correlated with the larger transferability score between two tasks.  
Fig. \ref{fig:minspantree}.a shows the task curriculum for the eight  tasks in Section 4.2. Furthermore, we did a similar experiments on  a collection of binary object classification tasks using the NUS-WIDE multi-label dataset \cite{nus-wide-civr09}  (Fig. \ref{fig:minspantree}.b). 
We set a threshold to find the most salient transfers and  the resulting curriculum is in line with human perception. 



%% file: conclusion.tex
\section{Conclusion}

\label{sec:conclusion}
In this paper, we presented H-score, an information theoretic approach to estimating the performance of features when transferred across classification tasks. Then we used it to define a notion of task transferability in multi-task transfer learning problems, that is both time and sample complexity efficient. 
Our transferability score  successfully predicted the performance for transfering features from ImageNet-1000 classification task to Cifar-100 task. 
Moreover, we showed how the transferability metric can be applied to a set of diverse computer vision and image-based recognition tasks using the Taskonomy and NUS-WIDE datasets. 
In future works, 
we will investigate properties of higher order transferability, developing more scalable algorithms that avoid computing the H-score of all task pairs.   
We also hope to design better task curriculum  for task transfer learning in practical applications. 
 

%% file: supplementary.tex
\section*{Supplementary Material} 
  \appendix 
  \renewcommand{\thesection}{S\arabic{section}}
  
        \setcounter{table}{0}
        \renewcommand{\thetable}{S\arabic{table}}%
        \setcounter{figure}{0}
        \renewcommand{\thefigure}{S\arabic{figure}}%
%
 \section{Derivation of Equation (4)}
 First, we need to introduce some additional notations. Let $X$, $x$, $\mathcal{X}$ and $P_{X}$ represent a random variable, a  value, the alphabet and the probability distribution respectively. $\sqrt{\mathrm{P}_{X}}$ denotes the vector with entries $\sqrt{P_{X}(x)}$ and $[\sqrt{\mathrm{P}_{X}}]\in \mathbb{R}^{|\X|\times|\X|}$ denotes the diagonal matrix of $\sqrt{\mathrm{P}_{X}}$. 
 For joint distribution $P_{YX}$, $\mathrm{P}_{YX}\in\R^{|\Y|\times|\X|}$  represents the probability matrix.   
 Given k feature functions $f_{i}:\mathcal{X} \rightarrow \mathbb{R}, i=1,...,k$, let $f(x)=[f_1 (x),...,f_k (x)]\in \mathbb{R}^{k}$ be the   feature vector of $x$, and $F=[f(x_1)^\mathrm{T},...,f(x_{|\mathcal{X}|})^\mathrm{T}]^\mathrm{T}\in \mathbb{R}^{|\X|\times k}$ be the feature matrix over all elements in $\X$.
 
The left hand side of Equation (4) can be expressed as
\begin{align}
||\tilde{B} \Phi (\Phi^\mathrm{T}\Phi)^{-\frac{1}{2}} ||_F^2 =& \textup{tr}\left( ( \Phi^\mathrm{T}\Phi)^{-\frac{1}{2}}\Phi ^{\mathrm{T}}\tilde{B}^{\mathrm{T}}\tilde{B} \Phi (\Phi^\mathrm{T}\Phi)^{-\frac{1}{2}} \right) \nonumber \\
=& \textup{tr}\left( ( \Phi^\mathrm{T}\Phi)^{-1}\Phi ^{\mathrm{T}} \tilde{B}^{\mathrm{T}}\tilde{B} \Phi \right)  \label{eq:form} 
\end{align}

\noindent 
Since any feature function can be centered by subtracting the mean, without the loss of generality, we assume $\E[f(X)]=0$. Using the one-to-one correspondence between $\Phi$ and $F$, i.e.  $\Phi=\left[\sqrt{\mathrm{P}_{X}}\right]F \in \mathbb{R}^{|\mathcal{X}|\times k}$,   we have
\begin{align}
\label{eq:normalization}
\Phi ^{\mathrm{T}}\Phi &= \left(\left[\sqrt{\mathrm{P}_{X}}\right]F\right)^{\mathrm{T}}\left(\left[\sqrt{\mathrm{P}_{X}}\right]F \right) \nonumber\\
&= \E[f(X)^{\mathrm{T}}f(X)] \nonumber\\
&= \textup{cov}(f(X)) 
\end{align}
The DTM matrix $\B$ introduced in Definition 1 can be written in matrix notation: \(
\tilde{B}= \left[ \sqrt{\mathrm{P} _{Y}} \right]^{-1}\mathrm{P}_{YX}\left[ \sqrt{\mathrm{P} _{X}} \right]^{-1}-\sqrt{\mathrm{P} _{Y}}\sqrt{\mathrm{P} _{X}}^{\mathrm{T}}.
\) Then we have,  
\begin{align*}
\tilde{B} \Phi=&\left(\left[\sqrt{\mathrm{P} _{Y}}\right]^{-1}\mathrm{P}_{YX}\left[\sqrt{\mathrm{P} _{X}}\right]^{-1}-\sqrt{\mathrm{P} _{Y}}\sqrt{\mathrm{P} _{X}}^{\mathrm{T}}\right)\cdot \\ &\left[\sqrt{\mathrm{P} _{X}}\right]F \\
=& \left[\sqrt{\mathrm{P} _{Y}}\right]\left(\left[\mathrm{P} _{Y}\right]^{-1}\mathrm{P}_{YX}F-\mathbf{1} \cdot \E[f(X)]^{\mathrm{T}} \right), 
\end{align*}
 where $\mathbf{1}$ is a column vector with all entries $1$ and length $|\Y|$. It follows that,
\begin{align}
\Phi ^{\mathrm{T}}\tilde{B}^{\mathrm{T}}\tilde{B}\Phi \nonumber 
= & \left(\left[ \mathrm{P} _{Y} \right]^{-1}\mathrm{P}_{YX}F-\mathbf{1} \cdot \E[f(X)]^{\mathrm{T}} \right)^{\mathrm{T}}\cdot \\ &\left[ \mathrm{P} _{Y} \right]\left(\left[\mathrm{P}_{Y}\right]^{-1}\mathrm{P}_{YX}F-\mathbf{1} \cdot \E[f(X)]^{\mathrm{T}}\right) \nonumber  \\
= & \E_{P_{Y}}\left[(\E[f(X)|Y]-\mathbf{1} \cdot \E[f(X)]^{\mathrm{T}})^{\mathrm{T}}\cdot \nonumber \right.\\
 & \left. (\E[f(X)|Y]-\mathbf{1} \cdot \E[f(X)]^{\mathrm{T}})\right]  \nonumber \\
= & \textup{cov}\left(\E[f(X)|Y]\right)  \label{eq:BXi}
\end{align}
By substituting (\ref{eq:normalization}) and (\ref{eq:BXi}) into (\ref{eq:form}), we have
\begin{equation*}
||\tilde{B} \Phi (\Phi^\mathrm{T}\Phi)^{-\frac{1}{2}} ||_F^2 = \tr\left(\textup{cov}(f(X))^{-1}\cov(\E[f(X)|Y])\right) \nonumber
\end{equation*}

\input{hypopthesis_testing} 
\section{Experiment Details}
\subsection{Experiment 4.1}
{\noindent\bf Experiment Setup. }The training data for the target task in this experiment consists of  $20,000$  images randomly sampled from the Cifar-$100$ dataset \cite{cifar}. It is further split 9:1 into a training set and a testing set.

We first extracted features of the Cifar-$100$ training images from five different layers (4a - 4f) of the ResNet-50 model pretrained on ImageNet-1000 \cite{imagenet}. Then we computed the H-score and the empirical transfer performance of each feature function for the Cifar-$100$ task.  To compute the empirical performance, we trained the transfer network  using stochastic gradient descent with batch size $20,000$ for 100 epochs. 

{\noindent\bf Result Discussion. }
\begin{figure}[htp]
\centering
\renewcommand{\arraystretch}{0.8}
\includegraphics[height=0.18\textheight]{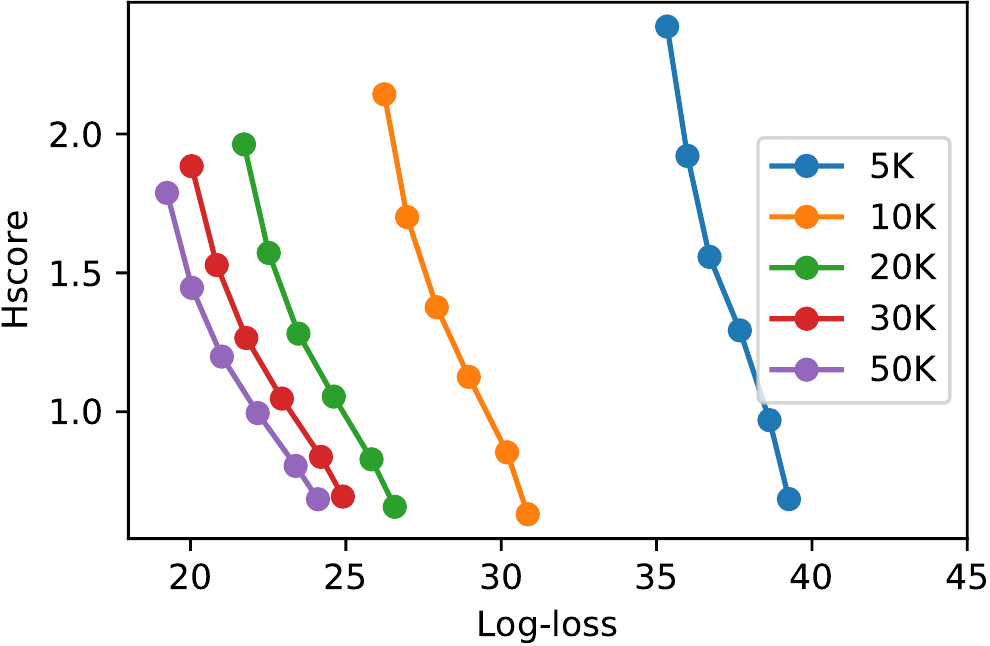} 
\caption{\label{fig:cifar}H-score and transferability vs. the empirical transfer performance measured by log-loss for different target sample size (5K-50K). } 
\end{figure}  
 As shown in Fig. 2.a of the main paper, transfer performance is better when an upper layer of the source networks is transferred. This could be due to the inherent similarity between  the target task and the source task, such that the optimal representation learned for one task can still be suitable for the other. 
For the experiment of selecting the best target task (Fig. 2.b), we used the same network as the former experiment to compute the empirical transfer performance with batch size 64 for 50 epochs. 

In addition, we validated  H-score under different target sample sizes between 5-50K. Fig.~\ref{fig:cifar} shows that target sample size does not affect the relationship between H-score and log-loss, which further demonstrates that the H-score computation is sample efficient.

\subsection{Experiment 4.2}

\noindent{\bf Data and Tasks. } The Taskonomy dataset \cite{zamir2018taskonomy} contains 4,000,000 images of indoor scenes collected from 600 buildings. Every image has annotations for $26$ computer vision tasks. For the transferability experiment, we  randomly sampled $20,000$ images as the target task training data, and selected eight supervised tasks, shown in Table \ref{tab:tasks}.

\begin{table}[h!] 
\setlength{\tabcolsep}{5pt}
  \begin{tabular}{lp{3.1cm}lp{1.3cm}}
    \toprule
 Tasks & Description & Output & Quantize-level \\
    \midrule
   Edge2D   &   2D edges detection    & images & 16 \\
    Edge3D  &  3D occlusion edges detection   & images & 16 \\
   Keypoint2D  &   2D keypoint detection  & images & 16 \\
  Keypoint3D  & 3D keypoint detection   & images & 16 \\
   Reshading & Image reshading   & images & 16 \\
   Depth   & Depth estimation     & images & 16 \\
    \midrule
Object Class. &    Object classification   & labels & none \\
Scene Class. & Scene classification    & labels & none \\
    \bottomrule
  \end{tabular}  
  \caption{\label{tab:tasks}Task descriptions}
\end{table}

\noindent{\bf Feature Extraction and Data Preprocessing. }\label{sssec:preprocess} 
For each task, \cite{zamir2018taskonomy} trained a fully supervised network with an encoder-decoder structure. When testing the transfer performance from source task  $\mathcal{T}_S$ to target task  $\mathcal{T}_T$, the encoder output of $\mathcal{T}_S$ is used for training the decoder of $\mathcal{T}_T$. 
For a fair comparison, we used the same trained encoders to extract source features. The output dimension of all encoders are   $16 \times 16 \times 8$ and we flattened the output into a vector of length $2048$. To reduce the computational complexity, we also resized the ground truth  images into $64\times 64$.

\begin{figure}
\centering
\includegraphics[width=0.4\textwidth]{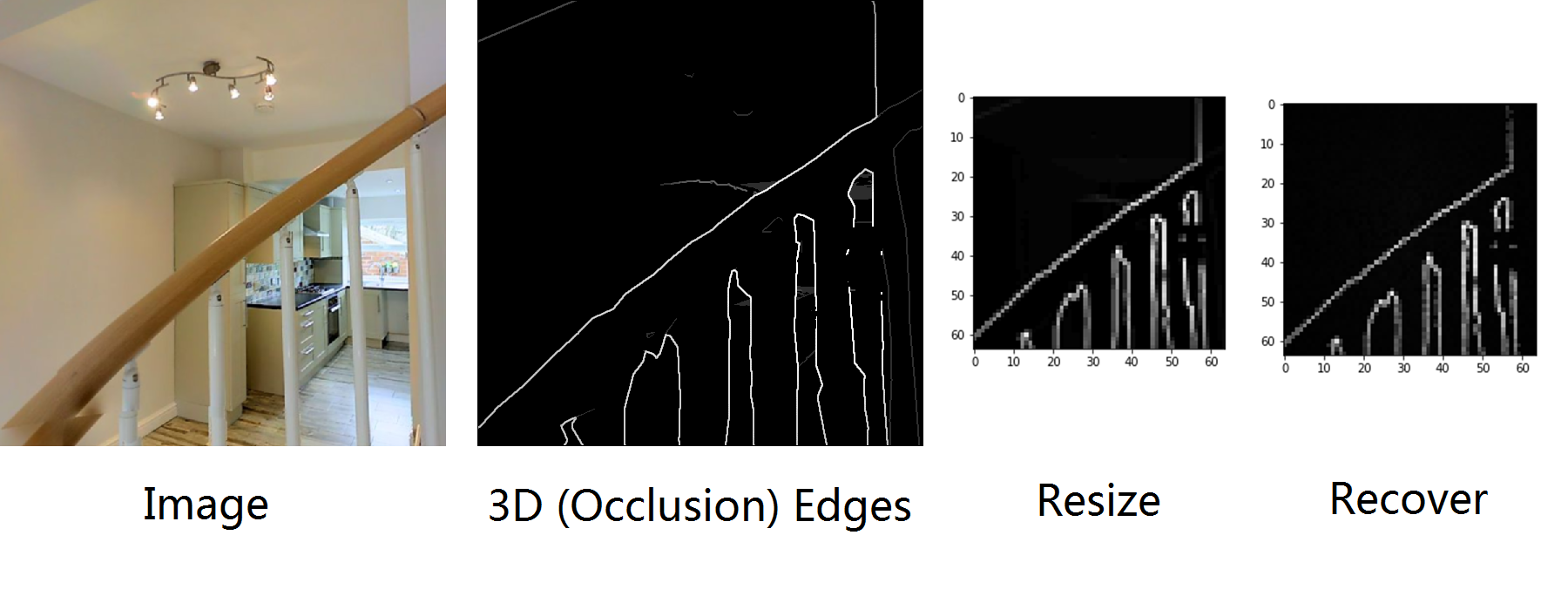}
\caption{Quantization. Recover is done with the centroid of corresponding cluster of each pixel.}
\label{fig:quantize}
\end{figure}

\begin{figure}[htp]
\centering
\includegraphics[width=0.4\textwidth]{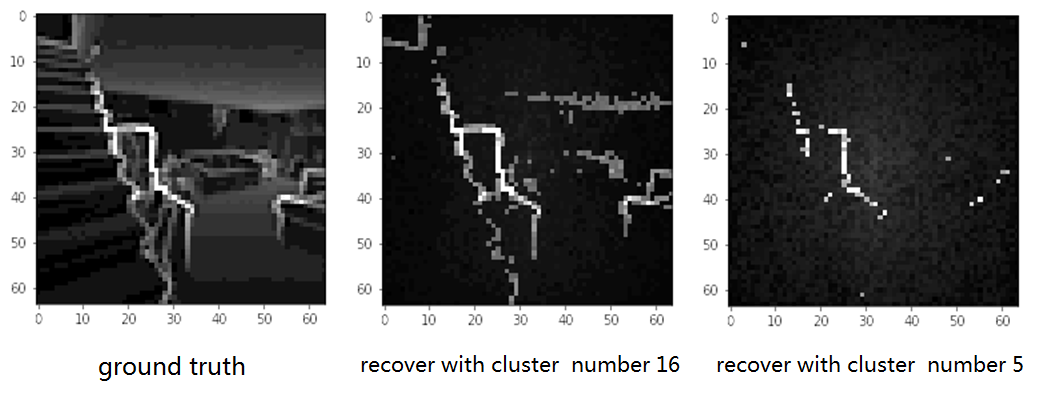}
\caption{Effect of quantization cluster size for 3D occlusion Edge detection. } 
\label{fig:recover}
\end{figure}

\noindent{\bf Label Quantization. }
  Fig.~\ref{fig:quantize} illustrates the resizing and quantization process of a pixel-to-pixel task label. 
During the quantization process, we are primarily concerned with two factors:   computational complexity and information loss. Too much information loss will lead to bad approximation of the original problem. On the other hand, having little information loss requires larger label space (cluster size) and higher computation cost.
To test the sensitivity of the cluster size, we use cluster centroids to recover the ground truth image pixel-by-pixel. The recovery results  for 3D occlusion edge detection is shown in Figure \ref{fig:recover}. When the cluster size is  $N=5$ (right), most detected edges in the ground truth image (left) are lost. We found that $N=16$ strikes a good balance between recoverability and computation cost. 
\\

\noindent{\bf Comparison of H-scores and Affinities.}
Table \ref{tab:value} presents the numerical values  of the  transferability and affinity scores between every pair of tasks,  with columns representing source tasks and rows representing target tasks. This table is in direct correspondence with the ranking matrices in Fig. 3 of the main paper. 
For each target task, the upper row shows our results while the lower one shows the results in \cite{zamir2018taskonomy}. Score values are included in parenteses. 
\input{tasko-tab}

\begin{figure}[htp]
\centering
\includegraphics[width=0.5\textwidth]{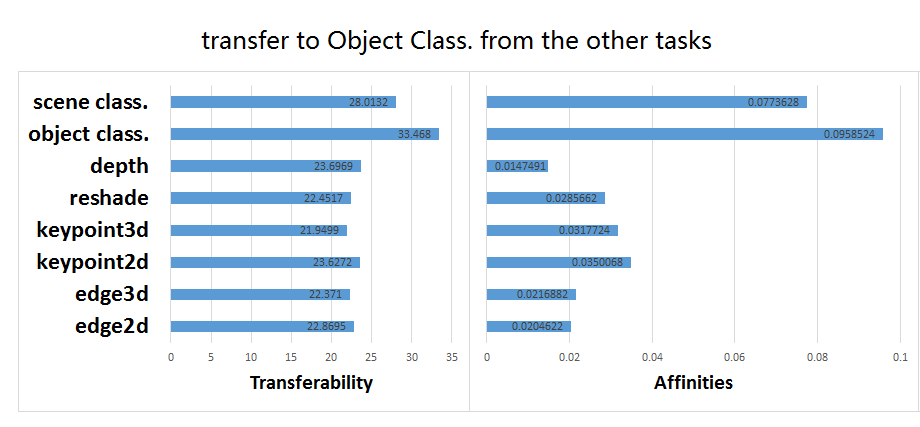}
\includegraphics[width=0.5\textwidth]{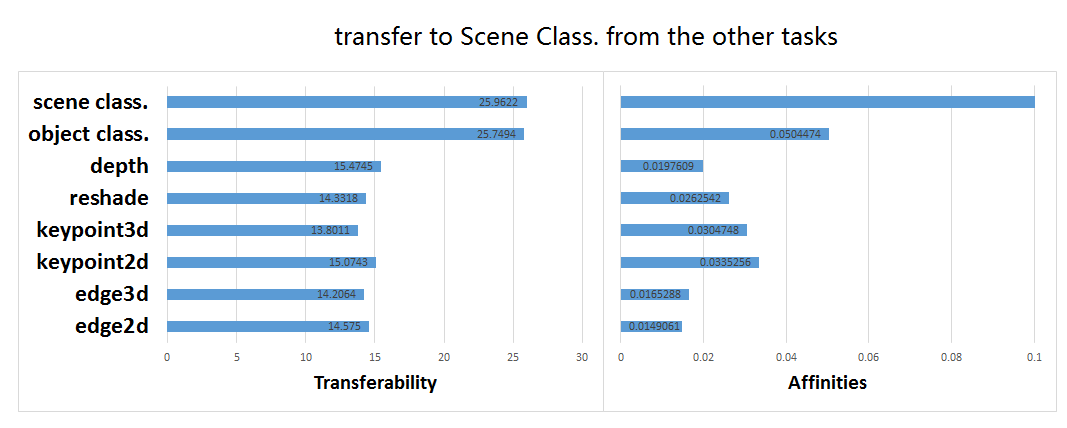}
\caption{Source task transferability ranking for classification tasks. For each target task,  the left figure shows H-score results, and the right figure shows task affinity results.}
\label{fig:class}
\end{figure}
\begin{figure}[htp]
\centering
\includegraphics[width=0.5\textwidth]{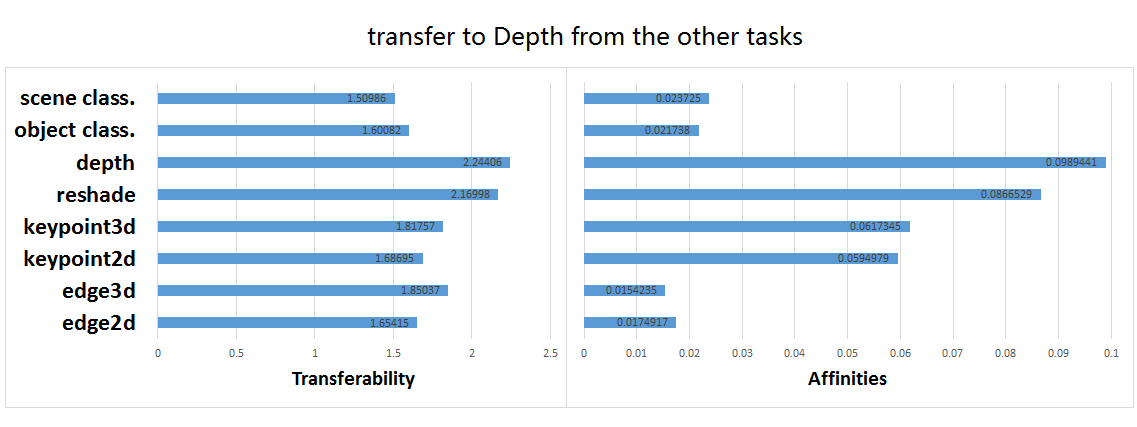}
\caption{Comparison between source task rankings for Depth. with H-score results on the left and affinity scores \cite{zamir2018taskonomy} on the right. The Top 3 transferable source tasks in both methods are the same: Depth, Image Reshading and 3D Occlusion Edges.} 
\label{fig:Depth}
\end{figure}

Here we present some detailed results on the comparison between H-score and the affinity score in \cite{zamir2018taskonomy} for pairwise transfer. 
The results of transferring from all tasks to the two classification tasks (Object Class. and Scene Class.) are shown in Figure \ref{fig:class}; The results of transferring to Depth is shown in \ref{fig:Depth}.  We can see in general, although affinity and transferability have totally different value ranges, they tend to agree on the order of the top few ranked tasks. 

\noindent {\bf Computing Efficiency.}
We ran the experiment on a workstation with  $3.40$ GHz $\times 8$ CPU and $16$ GB memory. Each pairwise H-score computation finished in less than one hour including preprocessing. 
\subsection{Experiment 4.3}
{\noindent\bf Data and Tasks. }  The NUS-WIDE dataset \cite{nus-wide-civr09} contains 161,789 web images for training and 107,859 images for evaluation. Its tag set consists of 81 concepts, among which we selected two  subsets, shown in Table \ref{tab:nuswide}. One subset contains 36 common concepts including scenes, animals and objects; The other subset contains only animal concepts. The NUS-WIDE dataset provides six types of low-level features. In this experiment, we used the 500 dimensional bag-of-words feature based on the SIFT descriptors.
\begin{table}[h]
    \centering
    \begin{tabular}{c|p{5cm}}
        Common concepts &  beach, birds, boats, bridge, buildings, cars, cat, clouds, dancing, dog, fish, flowers, garden, grass, house, lake, leaf, moon, mountain, ocean, person, plants, rainbow, rocks, running, sand, sky, sports, street, sun, swimmers, town, tree, vehicle, water, window   \\
        Animal concepts  & animal, bear, cat, cow, dog, elk, fox, horses, tiger, zebra 
    \end{tabular}
    \caption{Subsets of NUS-WIDE tag concepts used in Experiment 4.3}
    \label{tab:nuswide}
\end{table}

\begin{figure}[htp]
\centering
\renewcommand{\arraystretch}{0.8}
\includegraphics[height=0.2\textheight]{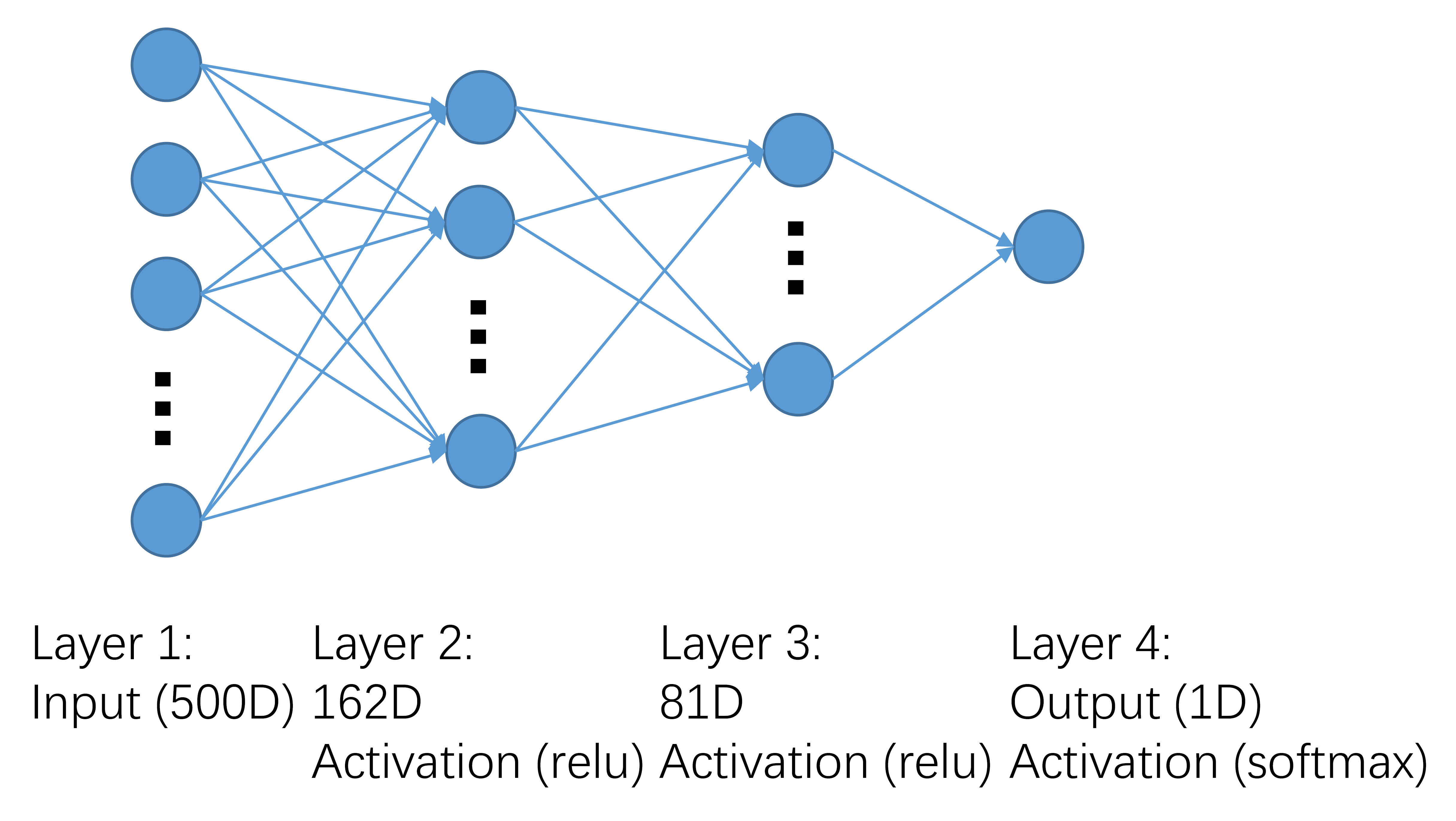} 
\caption{\label{fig:nn} Network structure for NUS-WIDE experiment.} 
\end{figure}

{\noindent\bf Feature Extraction and Data Preprocessing. } We consider the prediction of each concept as an unbalanced binary classification task and train a 4-layer fully-connected neural network (Fig. \ref{fig:nn}) with batch-size 2048 for 50 epochs. For an arbitrary target task, the layer 3  activation output of the source models are used to calculate the H-scores.

{\noindent\bf Task Transfer Curriculum. }
Given $n$ tasks $\T_1,\dots,\T_n$, first we compute the pairwise transferability matrix $M\in \R^{n\times n}$ using H-score,  where $M(i,j) = \frac{\H_{\T_j}(f_{\T_i})}{\H_{\T_j}(f_{\T_j})}$ for all $1\leq i,j\leq n$. We assume that the task-specific features are close to optimal, such that  $\H_{\T_j}(f_{\T_j})\approx \H_{\T_j}(f_{\T_{j}opt}) $. Then by Definition 3, $M(i,j)\approx \mathfrak{T}(\T_i,\T_j)$. Using the transferability matrix, we define an undirected graph $G$ over the tasks, where the edge weight between node $i$ and node $j$ is defined by 
\[
\small
W(i,j) = \begin{cases}
1-\max\{M(i,j), M(j,i)\} &\text{if }\parbox[t]{0.18\textwidth}{$\max\{M(i,j), M(j,i)\}$  $\geq \alpha$} \\
0 &\text{otherwise}
\end{cases}
\]
Parameter $\alpha$ defines the threshold to filter out less related task pairs, as the transferred representation in these cases contribute very little to the training of the target task.  
If $G$ is connected, the minimum spanning tree outputs a set of  task pairs that maximizes the total transferability with $n-1$ pairwise transfers. If $G$ is not connected, we have a minimum spanning forest that represents several task groups that can be learned independently. Finally, we recover the transfer directions on the minimum spanning tree from the transferability matrix.

For the subset with 36 common concepts in the NUS-WIDE experiment, we found that  when $\alpha=0$, i.e. no edges are filtered, 33 of the 35 edges in the resulting tree indicate transfers to concept 'sky' from other concepts. This phenomenon is reasonable in that sky and other concepts coexist in images with a high probability. To demonstrate the most significant task relationships, we set edge threshold  $\alpha$ to be the 2.3 percentile of all weights, resulting in the minimum spanning tree in Fig. 7 of the main paper. 
On the other hand, edge filtering is not needed for the Taskonomy tasks and the animal concepts in the NUS-WIDE experiment, since most tasks are transferable to a similar extent. Therefore we chose $\alpha=0$ in these cases.

\input{relatedworks}

 

%% file: hypopthesis_testing.tex
\section{Operational Meaning of H-Score}
In this section, we will show that H-score characterizes the asymptotic probability of error in the hypothesis testing context. We will start with  some background on error exponents from statistics, then explain how to estimate it using  information geometry. Finally, we will show how computing H-score   is in fact estimating the error exponent of a feature function on the sample data.

\subsection{Error Exponent and Hypothesis Testing}

Consider the binary hypothesis testing  problem over $m$ i.i.d. sampled observations $\{x^{(i)}\}_{i=1}^{m}\triangleq x^{m}$ with the following hypotheses:
$H_0: x^m \sim P_1$  or   $H_1: x^m \sim P_2$. 

 Let $P_{x^m}$ be the empirical distribution of the samples. The optimal test, i.e., the log likelihood ratio test $\log(T)=\log\frac{P_{1}(x^{m})}{P_{2}(x^{m})}$ can be stated in terms of information-theoretic quantities as follows:
\[m[D(P_{x^{m}}||P_{2}) -D(P_{x^{m}}||P_{1})]\underset{H_{1}}{\overset{H_{0}}{\gtrless}}\log T \]
where $D$ is the Kullback-Leibler (KL) divergence operator.
  
\begin{figure}[t]
\centering
\begin{tikzpicture}
\begin{axis}[
  no markers, domain=-6:5, samples=100,
  axis lines=none,
  every axis y label/.style={at=(current axis.above origin),anchor=south},
  every axis x label/.style={at=(current axis.right of origin),anchor=west},
  height=5cm, width=10cm,
  xtick=\empty, ytick=\empty,
  enlargelimits=false, clip=false, axis on top,
  grid = major
  ]

  \addplot [fill=cyan!20, draw=none, domain=-6:0.2] {gauss(1,1)} \closedcycle;
  \addplot [fill=red!20, draw=none, domain=0.2:5] {gauss(-2,1)} \closedcycle;
  \addplot [very thick,cyan!50!black] {gauss(-2,1)};
  \addplot [very thick,cyan!50!black] {gauss(1,1)};
  
   \node [fill=none] at (axis cs:-4,0.35){$H_{0}: P_{1}$};
   \draw[->](axis cs:-3.2,0.2)--(axis cs:-4,0.3);
   \node [fill=none] at (axis cs:3,0.35){$H_{1}: P_{2}$};
   \draw[->](axis cs:2.2,0.2)--(axis cs:3,0.3);
   \node [fill=none] at (axis cs:-2,0.1){$\beta=P_{2}(A)$};
   \draw[->](axis cs:-1,0.03)--(axis cs:-2,0.07);
   \node [fill=none] at (axis cs:1.5,0.1){$\alpha=P_{1}(A^{c})$};
    \draw[->](axis cs:0.3,0.017)--(axis cs:1.5,0.07);
  \draw [dashed](axis cs:0.2,0) -- (axis cs:0.2,0.4);
  \draw[->](axis cs:0.2,0)--(axis cs:0.2,-0.1);
   \draw [latex-latex](axis cs:0.2,-0.05) -- node [fill=white] {$A$: fail to reject $H_{0}$} (axis cs:-6,-0.05);
   \draw [latex-latex](axis cs:0.2,-0.05) -- node [fill=white] {$A^{c}$:reject $H_{0}$} (axis cs:5,-0.05);
   \node [fill=white] at (axis cs:0.2,-0.15) {$P_{x^m}:D(P_{x^m}\|P_{2})-D(P_{x^m}\|P_{1})=\frac{\log T}{m}$};
\end{axis}

\end{tikzpicture}
\caption{The binary hypothesis testing problem. The blue curves shows the probility density functions for $P_1$ and $P_2$. The rejection region $A^c$ and the acceptance region $A$ are highlighted in red and blue, respectively. The vertical line indicates the decision threshold.  }
\label{fig:ht} 
\end{figure}
Further, using Sannov's theorem, we have the asymptotic probability of type \uppercase\expandafter{\romannumeral1} error:
\[ \alpha=P_{1}(A^{c}) \approx 2^{-mD(P_{1}^{*}||P_{1})} \]
where $P_{1}^{*} = \argmin _{P \in A^{c}}D(P||P_{1})$ and $A^{c}(T) = \{x^{m}: D(P_{x^{m}}||P_{2})-D(P_{x^{m}}||P_{1}) < \frac{1}{m} \log T \}$ represents the rejection region. Similarly, the asymptotic probability of  type \uppercase\expandafter{\romannumeral2} error is
\[ \beta=P_{2}(A) \approx 2^{-mD(P_{2}^{*}||P_{2})}, \]
 where $P_{2}^{*} = \argmin _{P \in A} D(P||P_{2})$ and 
 $A = \{x^{m}:$\\
  $D(P_{x^{m}}||P_{2}) -D(P_{x^{m}}||P_{1}) > \frac{1}{m} \log T \}$ represents the acceptance region (See Figure~\ref{fig:ht}). Using the Bayesian approach for hypothesis testing,    the overall error probability  of the log likelihood ratio test is defined as:
  \[ 
  P_{e}^{(m)}=\alpha P_1+\beta P_2
  \] and the {\it best achievable exponent in the Bayesian probability of error} (a.k.a. {\it error exponent}) is defined as:
\[ E = \lim_{m\rightarrow \infty}\min_{A \subseteq \mathcal{X}^{m}} -\frac{1}{m} \log P_e^{(m)}\]
 Error exponent $E$ expresses the best rate at which  the error probability decays as sample size increases for a particular hypothesis testing problem. See \cite{Cover:1991} for more background information on error exponents and its related theorems.

 \subsection{Estimating Error Exponents}  
   Suppose $P_1$ and $P_2$ are sampled from the $\epsilon$-neighborhood  $\mathcal{N}_{\epsilon}(P_0) \triangleq \{ P | \sum_{x \in \mathcal{X}}\frac{(P(x)-P_0(x))^{2}}{P_0(x)} \leq \epsilon^{2}\} $    centered at a reference distributon $P_0$, and let $\phi_1,\phi_2\in \R^{|\X|}$ be vectors defined as: 
   \[
   \phi_i(x)\triangleq \frac{P_i(x)-P_0(x)}{\epsilon\sqrt{P_0(x)}} 
   \]      for $i=1,2$. The following lemma express the optimal error exponent $E$  using $\phi_1$ and $\phi_2$: 

\begin{lemma}
Under the local assumption defined earlier, the best achievable error exponent of the binary hypothesis testing problem with probabilities $P_1$ and $P_2$ is: 
 \[  E=
 \frac{\epsilon^2}{8}\left \| \phi _{1}-\phi_{2} \right \|^{2}+o(\epsilon^{2})\]
 where $\epsilon$ is a constant  \cite{makur2015efficient}. 
 \end{lemma}
 While the above lemma characterizes the asymptotic error probability distinguishing $P_1$ and $P_2$ based on the optimal decision function, most decision functions we learn from data are not optimal, as $P_1$ and $P_2$ are unknown.
Given sample data $x^m$ and an arbitrary feature function $f:\X\rightarrow\R$, which could be learned from a pre-trained model, the error exponent of the decision function based on $f$ is  reduced in a way defined by the following Lemma:
  
\begin{lemma}\label{lemma:mismatch}Given a zero-mean, unit variance feature function $f:\X\rightarrow \R$ and i.i.d. sampled data $x^m$, the error probability of a mismatched decision function of the form $l=\frac{1}{m}\sum_{i=1}^m (f(x^{(i)}))$ has an exponent   
\[
E_{f} = \frac{\epsilon^2}{8}\langle \xi, \phi _{1}-\phi_{2} \rangle ^{2}+o(\epsilon^{2})
\] where $\xi\in\R^{|\X|}$ is a vector with entries $\xi(x) = \sqrt{P_{0}(x)}f(x)$ \cite{makur2015efficient}.
\end{lemma} 


This lemma characterizes the  error probability of using a normalized feature of the input data to solve a learning task by a linear projection of this feature between the input and output domains. Note that normalizing features to zero-mean and unit variance results in an equivalent decision function with a different threshold value, thus we can apply Lemma 2   to any features without the loss of generality. Further, it's obvious   that  the reduced   exponent $E_f$ is maximized when $\xi = \phi_1-\phi_2$, and the optimal value is exactly the optimal error exponent $E$ in Lemma 1.
To estimate the reduced error exponent for multi-dimensional features,  we present the $k$-dimensional generalization of Lemma~\ref{lemma:mismatch} below: 
\begin{lemma}
\label{lemma:mismatchK}
Given $k$ normalized feature functions $f(x) = [f_1(x),\dots, f_k(x)]$, such that $\E[f_i(X)]=0$ for all $i$, and $\cov(f(X))=I$ , we define a $k$-d statistics of the form $l^{k}=(l_{1},...,l_{k})$ where $l_{i}=\frac{1}{m}\sum_{l=1}^{m}f_{i}(x^{(l)})$. Let $\xi_1,\dots,\xi_k$ be $k$ vectors  with entries  $\xi_i(x) = \sqrt{P_X(x)}f_i(x)$ , $0\leq i\leq k$. Then the error exponent of $l^k$ is
\begin{equation}\label{eq:Ehkd}
E^k_{f} =\sum_{i=1}^k E_{f_i} = \sum_{i=l}^k \frac{\epsilon^2}{8}\left \langle \xi_i, \phi _1-\phi_2 \right \rangle ^{2} + o(\epsilon ^{2})
\end{equation} 
\end{lemma}
 The proof of this Lemma can be found in \cite{acejournal}.

\subsection{H-score and Error Exponents}
Now we return to the binary classification problem. Using Lemma 3, we will show the linear relationship between H-score and error exponents.
\begin{thm}\label{thm:errexpokD}
 Given $P_{X|Y=0}, P_{X|Y=1} \in \mathcal{N}_{\epsilon}^{\mathcal{X}}(P_{0,X})$ and features $f$ such that $\E \left[ f(X)\right]=0$ and    $\E [f(X)f(X)^\mathrm{T}]= I$, there exists some constant $c$ independent of $f$ such that  $E_f^{k}=c\mathcal{H}(f)$.  
\end{thm}
\begin{proof}
By Lemma \ref{lemma:mismatchK}, the L.H.S. of the equation can be written as $E_f^k =c_0 \sum_{i=l}^k \left \langle \xi_i, \phi _1-\phi_2 \right \rangle ^{2}  $ for some constant $c_0$. It follows that
\begin{align*}
c_0 &\sum_{i=l}^k \left \langle \xi_i, \phi _1-\phi_2 \right \rangle ^{2} \\
=& c_0 \left(\left(\mathrm{P}_{X|Y=0}-\mathrm{P}_{X|Y=1}\right) ^{\mathrm{T}}F\right)\left(\left( \mathrm{P}_{X|Y=0}-\mathrm{P}_{X|Y=1}\right)^{\mathrm{T}}F\right)^{\mathrm{T}}  \\
=& c_0 \left(\E[f(X)|Y=0]-\E[f(X)|Y=1]\right)^{\mathrm{T}} \cdot \\ &\left(\E[f(X)|Y=0]-\E[f(X)|Y=1]\right) \\
=& c_0 \frac{P_Y(0)+P_Y(1)}{P_Y(0)P_Y(1)}  \left(\frac{P_{Y}(0)P_{Y}(1)+P_{Y}(1)^2}{P_{Y}(0)}\right) \cdot \\ &\mathbb{E}\left[f(X)|Y=1\right]^{\mathrm{T}}\mathbb{E}\left[f(X)|Y=1\right] \\
=& c~\tr\left(\cov(\E[f(X)|Y])\right)\\
=& c~\mathcal{H}(f)
\end{align*}
The last equation uses the fact $\cov(f(X))=I$.
 \end{proof}


%% file: tasko-tab.tex
\begin{table*}[htp]
  \caption{Transferability ranking comparison, between H-score's estimation and task affinity }
  \label{tab:value}
  \centering
   \scalebox{0.82}[0.82]{
  \begin{tabular}{p{1.8cm}llllllll}
    \toprule
	 Tasks & 2D Edges& 2D Keypoints& 3D Edges&  3D Keypoints& Reshading& Depth& Object Class.&  Scene Class.\\
    \midrule
    \multirow{2}*{2D Edges}& \cellcolor{yellow}1 (1.8216) &  \cellcolor{sblue}2 (1.7334) & 5 (1.5704)& 6 (1.5696)&4 (1.6146)& \cellcolor{sred}3 (1.6201)&7 (1.5097)& 8 (1.4402) \\
    \cmidrule(r){2-9}
	&\cellcolor{yellow}1 (0.0389)~~~	& \cellcolor{sblue}2 (0.0117) 	& 4 (5.8920e-5)	&\cellcolor{sred}3 (8.8011e-5)	& 7 (2.9001e-5)	& 8 (2.2110e-5)	& 5 (4.9141e-5)	& 6 (4.8720e-5)\\	
     \midrule
    \multirow{2}*{2D Keypoints}&\cellcolor{sblue}2 (1.6698) &  \cellcolor{yellow}1 (1.7859) & 7 (1.5248) & 5 (1.5287)& 4 (1.5481) & \cellcolor{sred}3 (1.5632)&  6 (1.5253)& 8 (1.4725) \\
    \cmidrule(r){2-9}
    & \cellcolor{sblue}2 (0.0002)	&\cellcolor{yellow}1 (0.0542) 	& 7 (7.7797e-5)	& 5 (8.1029e-5)	&6 (7.8464e-5)	& 8 (7.2724e-5)	& \cellcolor{sred}3 (0.0002)	& 4 (0.0001)  \\
    \midrule
    \multirow{2}*{3D Edges}&5 (1.4828) &4 (1.4910)&\cellcolor{sred}3 (1.5167)&7 (1.4701)& \cellcolor{sblue}2 (1.5405)& \cellcolor{yellow}1 (1.6739)& 8 (1.4644)& 6  (1.4730) \\
    \cmidrule(r){2-9}
&6 (0.0117)  	&7 (0.0108) 	&\cellcolor{yellow}1 (0.1179)  	& \cellcolor{sblue}2 (0.0734) 	&4 (0.0622) 	&\cellcolor{sred}3 (0.0636) &8 (0.0094)	&5 (0.0151)  \\
    \midrule
    \multirow{2}*{3D Keypoints}&6 (1.5375)& 5 (1.5466) & 4 (1.5910)& \cellcolor{sred}3 (1.6456)& \cellcolor{yellow}1 (1.7198)& \cellcolor{sblue}2 (1.7122)& 7 (1.4709)& 8 (1.4121) \\
    \cmidrule(r){2-9}
&5 (0.0141)& 6 (0.0136) 	&\cellcolor{sblue}2 (0.0531) 	&\cellcolor{yellow}1(0.1275)	& \cellcolor{sred}3 (0.0400)	&4 (0.0247)	&7 (0.0132) 	&8 (0.0121)  \\    
    \midrule
    \multirow{2}*{Reshading}&5 (1.5504)&6 (1.5426) & \cellcolor{sred}3 (1.8174)&4 (1.7990)& \cellcolor{yellow}1 (2.2339)& \cellcolor{sblue}2 (2.1200)& 7 (1.4774)& 8 (1.3804) \\
    \cmidrule(r){2-9}
 &6 (0.0147) 	&8 (0.0143) 	&\cellcolor{sblue}2 (0.0781) 	&4(0.0545) 	&\cellcolor{yellow}1 (0.1121)	&\cellcolor{sred}3 (0.0765)	&7 (0.0144) 	& 5 (0.0174) \\   
    \midrule
    \multirow{2}*{Depth}&6 (1.6542) & 5 (1.6870)& \cellcolor{sred}3 (1.8504)& 4 (1.8176)& \cellcolor{sblue}2 (2.1700)& \cellcolor{yellow}1 (2.2441) & 7 (1.6008) & 8 (1.5099) \\
    \cmidrule(r){2-9}
     &7 (0.0175) 	& 8 (0.0154) 	&\cellcolor{sred}3 (0.0595) 	&4 (0.0617) 	& \cellcolor{sblue}2 (0.0867) 	&\cellcolor{yellow}1 (0.0989) 	&6 (0.0217) 	&5 (0.0237) \\
    \midrule
    \multirow{2}*{Object Class.}&5 (22.866) &4 (23.627) & 7 (22.371) &  8 (21.950) & 6 (22.452) & \cellcolor{sred}3 (23.697) & \cellcolor{yellow}1 (33.468)& \cellcolor{sblue}2 (28.013) \\
    \cmidrule(r){2-9}
    &7 (0.0205) 	&6 (0.0217) 	& \cellcolor{sred}3 (0.0350)	&4 (0.0318) 	&5 (0.0286) 	&8 (0.0147) 	&\cellcolor{yellow}1 (0.0959) 	&  \cellcolor{sblue}2 (0.0774) \\
    \midrule
    \multirow{2}*{Scene Class.}& 5 (14.575) & 4 (15.074)& 7 (14.206)&  8 (13.801)& 6 (14.332)& \cellcolor{sred}3 (15.474)& \cellcolor{sblue}2 (25.750)&  \cellcolor{yellow}1 (25.962) \\
    \cmidrule(r){2-9}
& 8 (0.0149) 	&7 (0.0165) 	&\cellcolor{sred}3 (0.0335) 	&4 (0.0305) &5 (0.0263)  	&6 (0.0198) 	& \cellcolor{sblue}2 (0.0504)  & \cellcolor{yellow}1 (0.1474) \\
    \bottomrule
  \end{tabular}
  }
\end{table*}

%% file: relatedworks.tex
\section{Related Works}
\label{sec:relatedWorks}
{\bf Transfer learning. }Transfer learning can be devided into two  categories: 
 {\em domain adaptation}, where knowledge transfer is achieved by making representations learned from one input domain work on a different input domain, e.g. adapt models for RGB images to infrared images \cite{wang2018deep}; and {\em  task transfer learning}, where knowledge is transferred between different  tasks on the same input domain~\cite{torrey2010transfer}.  
 Our paper focus on the latter prolem. 

{\noindent\bf Empirical studies on transferability.} \cite{yosinski2014transferable} compared the transfer accuracy of features from different layers in a neural network between image classification tasks. 
A similar study was performed for NLP tasks by  \cite{conneau2017supervised}.    \cite{zamir2018taskonomy} determined the optimal transfer hierarchy over a collection of perceptual indoor scene understanidng tasks, while transferability was measured by a non-parameteric score called "task affinity" derived from neural network transfer losses coupled with an ordinal normalization scheme.

{\noindent\bf Task relatedness.  }  
One approach to define task relatedness is based on task generation.  Generalization bounds have been derived for multi-task learning \cite{baxter2000model},   learning-to-learn  \cite{maurer2009transfer} and life-long learning  \cite{Pentina2014A}. Although these studies show theoretical results on transferability, it is hard to infer from data whether  the assumptions are satisfied. Another approach is estimating task relatedness from data, either  explicitly \cite{bonilla2008multi,zhang2013heterogeneous} or implicitly   as a regularization term on the network weights \cite{xue2007multi,jacob2009clustered}.  
 Most works in  this category are limited to shallow ones in terms of the model parameters.  
 
{\noindent\bf Representation learning and evaluation. }Selecting optimal features for a given task is traditionally performed via feature subset  selection or feature weight learning. Subset selection chooses features with maximal relevance and minimal redundancy according to information theoretic or statistical criteria   \cite{peng2005feature,hall1999correlation}. The feature weight approach learns the task while regularizing feature weights with sparsity constraints, which is common in multi-task learning \cite{liao2006radial, argyriou2007multi}. In a different perspective,  \cite{huang2017information} consider  the universal feature selection problem, which finds the most informative features from data when the exact inference problem is unknown. When the target task is given, the universal feature is equivalent to the minimum error probability feature used in this work.

%% file: main.bbl
\begin{thebibliography}{10}

\bibitem{pratt1993}
Lorien~Y Pratt,
\newblock ``Discriminability-based transfer between neural networks,''
\newblock in {\em Advances in neural information processing systems}, 1993, pp.
  204--211.

\bibitem{donahue2014decaf}
Jeff Donahue, Yangqing Jia, Oriol Vinyals, Judy Hoffman, Ning Zhang, Eric
  Tzeng, and Trevor Darrell,
\newblock ``Decaf: A deep convolutional activation feature for generic visual
  recognition,''
\newblock in {\em International conference on machine learning}, 2014, pp.
  647--655.

\bibitem{7318461}
C.~K. Shie, C.~H. Chuang, C.~N. Chou, M.~H. Wu, and E.~Y. Chang,
\newblock ``Transfer representation learning for medical image analysis,''
\newblock in {\em 2015 37th Annual International Conference of the IEEE
  Engineering in Medicine and Biology Society (EMBC)}, Aug 2015, pp. 711--714.

\bibitem{gaodeep}
Yuqing Gao and Khalid~M Mosalam,
\newblock ``Deep transfer learning for image-based structural damage
  recognition,''
\newblock {\em Computer-Aided Civil and Infrastructure Engineering}.

\bibitem{yosinski2014transferable}
Jason Yosinski, Jeff Clune, Yoshua Bengio, and Hod Lipson,
\newblock ``How transferable are features in deep neural networks?,''
\newblock in {\em Advances in neural information processing systems}, 2014, pp.
  3320--3328.

\bibitem{zamir2018taskonomy}
Amir Zamir, Alexander Sax, William Shen, Leonidas Guibas, Jitendra Malik, and
  Silvio Savarese,
\newblock ``Taskonomy: Disentangling task transfer learning,''
\newblock {\em Computer Vision and Pattern Recognition (CVPR)}, 2018.

\bibitem{conneau2017supervised}
Alexis Conneau, Douwe Kiela, Holger Schwenk, Loic Barrault, and Antoine Bordes,
\newblock ``Supervised learning of universal sentence representations from
  natural language inference data,''
\newblock {\em arXiv preprint arXiv:1705.02364}, 2017.

\bibitem{baxter2000model}
Jonathan Baxter,
\newblock ``A model of inductive bias learning,''
\newblock {\em Journal of artificial intelligence research}, vol. 12, pp.
  149--198, 2000.

\bibitem{maurer2009transfer}
Andreas Maurer,
\newblock ``Transfer bounds for linear feature learning,''
\newblock {\em Machine learning}, vol. 75, no. 3, pp. 327--350, 2009.

\bibitem{Pentina2014A}
Anastasia Pentina and Christoph~H. Lampert,
\newblock ``A pac-bayesian bound for lifelong learning,''
\newblock in {\em International Conference on International Conference on
  Machine Learning}, 2014, pp. II--991.

\bibitem{ben2003exploiting}
Shai Ben-David, Reba Schuller, et~al.,
\newblock ``Exploiting task relatedness for multiple task learning,''
\newblock {\em Lecture notes in computer science}, pp. 567--580, 2003.

\bibitem{acejournal}
Shao-Lun Huang, Anuran Makur, Gregory~W. Wornell, and Lizhong Zheng,
\newblock ``On universal features for high-dimensional learning and
  inference,'' \url{http://allegro.mit.edu/~gww/unifeatures}, 2019.

\bibitem{huang2017information}
Shao-Lun Huang, Anuran Makur, Lizhong Zheng, and Gregory~W Wornell,
\newblock ``An information-theoretic approach to universal feature selection in
  high-dimensional inference,''
\newblock in {\em Information Theory (ISIT), 2017 IEEE International Symposium
  on}. IEEE, 2017, pp. 1336--1340.

\bibitem{breiman1985estimating}
Leo Breiman and Jerome~H Friedman,
\newblock ``Estimating optimal transformations for multiple regression and
  correlation,''
\newblock {\em Journal of the American statistical Association}, vol. 80, no.
  391, pp. 580--598, 1985.

\bibitem{wang2019}
Lichen Wang, Jiaxiang Wu, Shao-Lun Huang, Lizhong Zheng, Xiangxiang Xu, Lin
  Zhang, and Junzhou Huang,
\newblock ``An efficient approach to informative feature extraction from
  multimodal data,''
\newblock {\em AAAI}, 2019.

\bibitem{makur2015efficient}
Anuran Makur, Fabi{\'a}n Kozynski, Shao-Lun Huang, and Lizhong Zheng,
\newblock ``An efficient algorithm for information decomposition and
  extraction,''
\newblock in {\em Communication, Control, and Computing (Allerton), 2015 53rd
  Annual Allerton Conference on}. IEEE, 2015, pp. 972--979.

\bibitem{nus-wide-civr09}
Tat-Seng Chua, Jinhui Tang, Richang Hong, Haojie Li, Zhiping Luo, and Yan-Tao
  Zheng,
\newblock ``Nus-wide: A real-world web image database from national university
  of singapore,''
\newblock in {\em Proc. of ACM Conf. on Image and Video Retrieval (CIVR'09)},
  Santorini, Greece., July 8-10, 2009.

\bibitem{Cover:1991}
Thomas~M. Cover and Joy~A. Thomas,
\newblock {\em Elements of Information Theory},
\newblock Wiley-Interscience, New York, NY, USA, 1991.

\bibitem{cifar}
Alex Krizhevsky and Geoffrey Hinton,
\newblock ``Learning multiple layers of features from tiny images,''
\newblock 2009.

\bibitem{imagenet}
Alex Krizhevsky, Ilya Sutskever, and Geoffrey~E Hinton,
\newblock ``Imagenet classification with deep convolutional neural networks,''
\newblock in {\em Advances in neural information processing systems}, 2012, pp.
  1097--1105.

\bibitem{wang2018deep}
Mei Wang and Weihong Deng,
\newblock ``Deep visual domain adaptation: A survey,''
\newblock {\em Neurocomputing}, 2018.

\bibitem{torrey2010transfer}
Lisa Torrey and Jude Shavlik,
\newblock ``Transfer learning,''
\newblock in {\em Handbook of research on machine learning applications and
  trends: algorithms, methods, and techniques}, pp. 242--264. IGI Global, 2010.

\bibitem{bonilla2008multi}
Edwin~V Bonilla, Kian~M Chai, and Christopher Williams,
\newblock ``Multi-task gaussian process prediction,''
\newblock in {\em Advances in neural information processing systems}, 2008, pp.
  153--160.

\bibitem{zhang2013heterogeneous}
Yu~Zhang,
\newblock ``Heterogeneous-neighborhood-based multi-task local learning
  algorithms,''
\newblock in {\em Advances in neural information processing systems}, 2013, pp.
  1896--1904.

\bibitem{xue2007multi}
Ya~Xue, Xuejun Liao, Lawrence Carin, and Balaji Krishnapuram,
\newblock ``Multi-task learning for classification with dirichlet process
  priors,''
\newblock {\em Journal of Machine Learning Research}, vol. 8, no. Jan, pp.
  35--63, 2007.

\bibitem{jacob2009clustered}
Laurent Jacob, Jean-philippe Vert, and Francis~R Bach,
\newblock ``Clustered multi-task learning: A convex formulation,''
\newblock in {\em Advances in neural information processing systems}, 2009, pp.
  745--752.

\bibitem{peng2005feature}
Hanchuan Peng, Fuhui Long, and Chris Ding,
\newblock ``Feature selection based on mutual information criteria of
  max-dependency, max-relevance, and min-redundancy,''
\newblock {\em IEEE Transactions on pattern analysis and machine intelligence},
  vol. 27, no. 8, pp. 1226--1238, 2005.

\bibitem{hall1999correlation}
Mark~Andrew Hall,
\newblock ``Correlation-based feature selection for machine learning,''
\newblock 1999.

\bibitem{liao2006radial}
Xuejun Liao and Lawrence Carin,
\newblock ``Radial basis function network for multi-task learning,''
\newblock in {\em Advances in Neural Information Processing Systems}, 2006, pp.
  792--802.

\bibitem{argyriou2007multi}
Andreas Argyriou, Theodoros Evgeniou, and Massimiliano Pontil,
\newblock ``Multi-task feature learning,''
\newblock in {\em Advances in neural information processing systems}, 2007, pp.
  41--48.

\end{thebibliography}
